\documentclass[11pt]{article}

\usepackage[final]{acl}

\usepackage{times}
\usepackage{latexsym}

\usepackage[T1]{fontenc}   

\usepackage[utf8]{inputenc}

\usepackage{microtype}

\usepackage{inconsolata}

\usepackage{graphicx}

\usepackage{booktabs}
\usepackage{xcolor,colortbl}
\usepackage{relsize}
\usepackage{xspace}

\usepackage{listings}
\usepackage{soul}    
\DeclareRobustCommand{\hlc}[2][yellow!35]{%
  {\sethlcolor{#1}\hl{#2}}%
}

\lstdefinestyle{plaincode}{
  basicstyle=\ttfamily\small\color{black},
  backgroundcolor=\color{white},
  columns=fullflexible,
  keepspaces=true,
  showstringspaces=false,
  frame=single,
  framerule=0.4pt,
  rulecolor=\color{black!30},
  escapeinside={(*@}{@*)},
  breaklines=true,            
  breakatwhitespace=false,    
  breakindent=1em,            
  postbreak=\mbox{\textcolor{gray}{$\hookrightarrow$}\space} 
}

\definecolor{hlAmber}{HTML}{FFECB3}   
\definecolor{hlGreen}{HTML}{C8E6C9}   
\definecolor{hlBlue}{HTML}{BBDEFB}    
\definecolor{hlLavender}{HTML}{D1C4E9}
\definecolor{hlRed}{HTML}{FFCDD2}     

\usepackage[dvipsnames]{xcolor}

\newif\ifshowcomments
\showcommentstrue               

\newcommand{\defineauthor}[3]{%
  \expandafter\newcommand\csname #1\endcsname[1]{%
    \ifshowcomments\textcolor{#2}{[#3: ##1]}\fi}}

\defineauthor{taylor}{RoyalBlue}{Taylor}

\DeclareRobustCommand{\systemname}{Opt-ICL\xspace}
\DeclareRobustCommand{\method}{Spectrum Tuning\xspace}
\DeclareRobustCommand{\methodshort}{SpecT\xspace}

\title{Opt-ICL at LeWiDi-2025: \\ Maximizing In-Context Signal from Rater Examples via Meta-Learning}

\author{Taylor Sorensen \\ Department of Computer Science \\  University of Washington \\ \texttt{tsor13@cs.washington.edu}
        \And
        Yejin Choi \\ Department of Computer Science \\ Stanford University \\ \texttt{yejinc@stanford.edu}}

\begin{document}
\maketitle

\begin{abstract}

Many natural language processing (NLP) tasks involve subjectivity, ambiguity, or legitimate disagreement between annotators. 
In this paper, we outline our system for modeling human variation. Our system leverages language models' (LLMs) in-context learning abilities, along with a two-step meta-learning training procedure for
1) post-training on many datasets requiring in-context learning
and 2)
specializing the model via in-context meta-learning to the particular data distribution of interest.
We also evaluate the performance of our system submission to the Learning With Disagreements (LeWiDi) competition, where it was the overall winner on both tasks.
Additionally, we perform an ablation study to measure the importance of each system component. We find that including rater examples in-context is crucial for our system's performance, dataset-specific fine-tuning is helpful on the larger datasets,
post-training on other in-context datasets is helpful on one of the competition datasets,
and that performance improves with model scale.

\end{abstract}

\section{Introduction}
\label{sec:intro}

\begin{table*}[htbp]
\small
\smaller[0]
\centering
\begin{tabular}{@{}lccccc@{}}
\toprule
\rowcolor{gray!20} \textsc{Perspectivist Task} & MP & CSC & Par & VEN & Average Rank \\
\rowcolor{gray!20} & (error rate $\downarrow$) & (abs. dist. $\downarrow$) & (abs. dist. $\downarrow$) & (error rate $\downarrow$) & \\
\midrule
Ours & \textbf{.289 (1)} & \textbf{.156 (1)} & \underline{.119 (2)} & \underline{.270 (2)} & - \\
Best other team & \underline{.300 (2)} & \underline{.172 (2)} & \textbf{.080 (1)} & \textbf{.124 (1)} & - \\
Most frequent label baseline & .316 & .239 & .362 & .345 & - \\
Random label baseline & .499 & .352 & .367 & .497 & - \\
\addlinespace
Ours (rank) & \textbf{1} & \textbf{1} & \underline{2 (2-way tie)} & \underline{2 (4-way tie)} & \textbf{1.5 (1)} \\
Best other team (name) & \underline{DeMeVa (2)} & \underline{DeMeVa (2)} & \textbf{twinther (1)} & \textbf{twinther (1)} & \underline{DeMeVa 2 (2)} \\
\midrule
\rowcolor{gray!20} \textsc{Soft Task} & MP & CSC & Par & VEN & Average Rank \\
\rowcolor{gray!20} & (Manh. dist. $\downarrow$) & (Wass. dist. $\downarrow$) & (Wass. dist. $\downarrow$) & (Manh. dist. $\downarrow$) & \\
\midrule
Ours & \textbf{.422 (1)} & \textbf{.746 (1)} & \textbf{1.20 (1)} & .449 (3) & - \\
Best other team & \textbf{.428 (1)} & \textbf{.792 (1)} & \textbf{.983 (1)} & \textbf{.233 (1)} & - \\
Most frequent label baseline & .518 & 1.17 & 3.23 & .595 & - \\
Random label baseline & .687 & 1.54 & 3.35 & .676 & - \\
\addlinespace
Ours (rank) & \textbf{1 (2-way tie)} & \textbf{1 (2-way tie)} & \textbf{1 (3-way tie)} & 3 (3-way tie) & \textbf{1.5 (1)} \\
Best other team (name) & \textbf{PromotionGo (1)} & \textbf{DeMeVa (1)} & \textbf{twinther (1)} & \textbf{twinther (1)} & \underline{DeMeVa 2.75 (2)} \\
\bottomrule
\end{tabular}
\caption{Competition final results. Our system had an average rank of 1.5 on both the perspectivist and soft tasks, and was the \textbf{overall winner for both tasks}. First place result bolded, second place underlined for each dataset. The competition organizers determined ties by a two-sided Wilcoxon signed-rank test with the rank leader on item-level scores, failing to reject a difference above $\alpha=.05$ (see Appendix \ref{app:ties}).}
\label{tab:results}
\end{table*}

Natural language processing (NLP) evaluations typically assume that there is a single correct answer (a.k.a., ``ground truth") and view annotator disagreement as a source of \textit{noise} to be eliminated, generally attributing rating variation to poor instructions, incomplete task specification, or noisy data.
However, oftentimes annotator disagreement can be a useful \textit{signal} of subjectivity, ambiguity, or multiple reasonable interpretations \cite{Aroyo_Welty_2015}.
Properly integrating this disagreement can be important for robustness, uncertainty calibration, and representing multiple viewpoints.
To address this, more and more have argued for focusing on methods for integrating human variation into evaluation and modeling \cite{basile-etal-2021-need, Gordon_2022}, including annotations from people from diverse backgrounds \cite{kirk2024prismalignmentdatasetparticipatory, aroyo2023dicesdatasetdiversityconversational}, and aligning AI systems with pluralistic values \cite{sorensen2024roadmappluralisticalignment}.

In order to inspire work towards these goals, the Learning With Disagreements (LeWiDi) competition \cite{lewidi} consists of four datasets across two tasks for modeling disagreement: one task for predicting how a particular annotator's ratings (``perspectivist" task) and one for predicting the distribution of labels that a pool of annotators gave (``soft label" task).
In this system paper, we outline our system submission.

Our system (\systemname, for Optimizing In-Context Learning) takes a fully perspectivist approach, trying to predict how an individual annotator rated each instance and then aggregating individual predictions into a distribution for the soft task.
It primarily leverages LLMs' in-context learning ability \cite{brown2020languagemodelsfewshotlearners, xie2022explanationincontextlearningimplicit}, including an annotator's train ratings directly in-context at inference time. On top of a pre-trained autoregressive language model, we additionally perform two steps of training:
post-training in order to enhance the models' in-context learning abilities and teach a unified prompt format (or, Spectrum Tuning, see \citealt{sorensenmethod}),
and dataset-specific fine-tuning. Both training steps can be seen as forms of meta-learning \cite{vanschoren2018metalearningsurvey, min2022metaicllearninglearncontext}, where the model is tasked with learning how best to fit to the in-context rater examples.

Our main contributions include:
our proposed system for modeling disagreement (\S \ref{sec:system}), which was \textbf{the overall winner on both competition tasks}, and
an ablation study outlining the effect of each system component (\S \ref{sec:results}).

In particular, we find that:
\begin{itemize}
    \item Including rater examples in-context is crucial for performance;
    \item Dataset-specific fine-tuning is helpful on larger datasets;
    \item Post-training on other in-context perspectivist datasets (\method) significantly helped on one dataset;
    \item Performance scales with model size, but size alone does not compensate for dataset-specific training.
\end{itemize}

\section{Background and Task Summary}
\label{sec:background}

The Learning With Disagreements competition (LeWiDi, \citealt{lewidi}) aims to evaluate machine learning systems' ability to engage with and model human variation. The competition spans four datasets which contain subjective judgments where raters may disagree:

\begin{itemize}
    \item the \underline{MultiPIco dataset (MP)} \cite{casola-etal-2024-multipico}, in which workers label whether or not a short exchange from Twitter/Reddit is ironic (binary);
    \item  the \underline{Conversational Sarcasm Corpus (CSC)} \cite{jang-frassinelli-2024-generalizable}, involving a 1-6 Likert scale for rating the level of sarcasm of a response given a context (6-way classification);
    \item A \underline{paraphrase detection dataset (Par)} (as of yet unpublished, shared by conference organizers) from Quora Question Pairs where annotators rate how strongly the questions are paraphrases of each other on a Likert scale from -5 to 5, along with an explanation (11-way classification); and
    \item the \underline{VariErrNLI (VEN)} dataset \cite{weber-genzel-etal-2024-varierr}, on which annotators reannotate premise/hypothesis pairs for entailment. Annotators could assign one or more labels from entailment, neutral, and contradiction and provide an explanation (3 binary classifications, with at least one positive label).
\end{itemize}
In addition, some basic demographic information is provided about annotators for all datasets.

\begin{table}[htbp]
\small
\centering
\begin{tabular}{@{}lcccc@{}}
\toprule
\rowcolor{gray!20} \textsc{Train Split} & MP & CSC & Par & VEN \\
\midrule
\# Ratings & 60,471 & 25,574 & 1,600 & 1,505 \\
\# Instances & 12,017 & 5,628 & 400 & 388 \\
\# Annotators & 506 & 872 & 4 & 4 \\
\# Mean Rat./Ann. & 119.5 & 29.4 & 400 & 360.8 \\
\# Min Rat./Ann. & 10 & 21 & 400 & 348 \\
\# Max Rat./Ann. & 147 & 38 & 400 & 373 \\
\addlinespace
\rowcolor{gray!20} \textsc{Dev Split} & MP & CSC & Par & VEN \\
\midrule
\# Ratings & 15,178 & 3,186 & 200 & 187 \\
\# Instances & 3,005 & 704 & 50 & 50 \\
\# Annotators & 506 & 850 & 4 & 4 \\
\addlinespace
\rowcolor{gray!20} \textsc{Test Split} & MP & CSC & Par & VEN \\
\midrule
\# Ratings & 18,693 & 3,224 & 200 & 199 \\
\# Instances & 3,756 & 704 & 50 & 50 \\
\# Annotators & 506 & 860 & 4 & 4 \\
\bottomrule
\end{tabular}
\caption{Dataset statistics across train, dev, and test splits for the four LeWiDi datasets. MP and CSC are much larger across the total number of ratings and the number of annotators.}
\label{tab:dataset_stats}
\end{table}

For dataset statistics, see Table \ref{tab:dataset_stats}. Notably, MP and CSC are much larger datasets than Par and VEN: the MP/CSC train data contains 50k/25k ratings from 506/872 annotators, while Par/VEN contain 1.6k/1.5k ratings from 4/4 annotators respectively.

Using these datasets, the competition constitutes two tasks: a \textit{``soft labeling" task}, where the goal is to predict a probability distribution over possible labels that best match the human annotator label distribution and a \textit{``perspectivist" task}, where the goal is to take on the perspective on an individual annotator and predict that particular annotator's label given prior demonstrations from that rater and (optionally) some demographic information.

For scoring submissions, the two binary datasets (MP/VEN) evaluate the soft task with Manhattan distance and the perspectivist task with error rate. The two Likert scale datasets (CSC/Par) are evaluated using Wasserstein distance for the soft task and absolute distance for the perspectivist task.

For additional information on the competition setup, please refer to \citet{lewidi}.

\section{System Overview}
\label{sec:system}

Our system consists of three components:
\begin{enumerate}
    \item \method (or \methodshort,  \citealt{sorensenmethod}): Post-training an autoregressive large language model (LLM) on a collection of datasets with human variation, stochasticity, or epistemic uncertainty;
    \item Dataset-specific fine-tuning on in-context demonstrations from each rater; and
    \item Inference with in-context annotator information and training demonstrations.
\end{enumerate}

Specifically, our system uses the \texttt{google/gemma-3-12b-pt} \cite{gemmateam2025gemma3technicalreport} language model.

\subsection{Prompt Structure}
Our method depends on LLMs' ability to do in-context learning \cite{brown2020languagemodelsfewshotlearners, xie2022explanationincontextlearningimplicit}. We adopt the prompting structure from \citet{sorensenmethod}, which has three components: a description (including a task description/any annotator demographics), inputs (the instance to rate), and outputs (the given rating). For example, here is a prompt from Par:

\begin{lstlisting}[style=plaincode]
Given a pair of questions from Quora Question Pairs (QQP), assign a Likert scale score from -5 to 5 indicating how strongly the questions are paraphrases of one another, and provide a short explanation for your score.
Annotator demographics: annotator_id: Ann1; Gender: Male; Age: 26; Nationality: Chinese; Education: master student
{"question1": "What are some things new employees should know going into their first day at Exact Sciences?", "question2": "What are some things new employees should know going into their first day at Garmin?", "lang": "en"}
<start_of_turn>(*@\hlc[hlAmber]{\{"paraphrase\_rating": -1, "explanation": "The companies are different."\}<end\_of\_turn>}@*)
{"question1": "Who are the everyday heroes and heroines of life?", "question2": "What was everyday life like under Nazi rule?", "lang": "en"}
<start_of_turn>(*@\hlc[hlGreen]{\{"paraphrase\_rating": -5, "explanation": "Q1 asks about everyday heroes and heroines. Q2 is aobut ever
yday life under nazi rule"\}<end\_of\_turn>}@*)
{"question1": "What does 'sandiaga' mean?", "question2": "What does \u064a\u0639\u0646\u064a mean?", "lang": "en"}
<start_of_turn>(*@\hlc[hlBlue]{\{"paraphrase\_rating":}...
\end{lstlisting}

The output of interest (in this case, a paraphrase rating and explanation) is wrapped in special tokens \texttt{<start\_of\_turn>/<end\_of\_turn>}.
While the LeWiDi competition evaluates only a systems' ability to predict the Likert/binary score, we include all rating data in the prompt (including the explanations) with the reasoning that 1) the rater's stated reasoning may contain predictive information for new examples and 2) training on the rating and the explanation concurrently may be helpful.

When predicting how a given rater may respond to a particular instance (e.g., the ``perspectivist" approach), we include their demographics at the beginning of the prompt, put as many example train ratings as will fit into context, and append the instance to evaluate at the end of the context.

Throughout the paper, we use a maximum context length of 3,000 tokens. With this limit, we are able to fit about 16 in-context examples for MP, 29 for CSC, 35 for Par, and 29 for VEN (See Table \ref{tab:prompt_stats}).

\begin{table}[htbp]
\small
\centering
\begin{tabular}{@{}lcccc@{}}
\toprule
\rowcolor{gray!20} \textsc{Dataset} & MP & CSC & Par & VEN \\
\midrule
\multicolumn{5}{c}{\textbf{In-Context Examples per Rater Prompt}} \\
\midrule
Mean & 15.8 & 28.6 & 35.0 & 29.1 \\
Min & 1 & 21 & 32 & 27 \\
Max & 32 & 37 & 41 & 31 \\
\midrule
\multicolumn{5}{c}{\textbf{Prompt Length (tokens)}} \\
\midrule
Mean & 2,542.1 & 2,492.1 & 2,717.1 & 2,707.3 \\
Min & 182 & 1,647 & 2,688 & 2,649 \\
Max & 2,798 & 2,776 & 2,757 & 2,769 \\
\addlinespace
\bottomrule
\end{tabular}
\caption{Prompt length and number of in-context examples used during inference across datasets.}
\label{tab:prompt_stats}
\end{table}

\begin{table*}[t!bp]
\small
\centering
\begin{tabular}{@{}lcccc@{}}
\rowcolor{gray!20} \textsc{Perspectivist Task} & MP & CSC & Par & VEN \\
\rowcolor{gray!20} & (error rate $\downarrow$) & (abs. dist. $\downarrow$) & (abs. dist. $\downarrow$) & (error rate $\downarrow$) \\
\midrule
\systemname (\methodshort + SFT + Demographics + ICL) & \textbf{.289 (1)} & \textbf{.156 (1)} & \textbf{.119 (1)} & \textbf{.270 (1)} \\
\textit{Prompt ablations} & & & & \\
\quad no demographics & \underline{.295 (2)} & \textbf{.156 (1)} & \textbf{.122 (1)} & \textbf{.268 (1)} \\
\quad no many-shot ICL (one example)          & .305 (3) & .185 (3) & .216 (3) & \underline{.321 (2)} \\
\textit{Training Ablations} & & & & \\
\quad no SFT          & .316 (4) & .191 (3) & \textbf{.123 (1)} & \textbf{.257 (1)} \\
\quad no \methodshort          & .303 (3) & \textbf{.157 (1)} & \textbf{.120 (1)} & \textbf{.247 (1)} \\
\quad no SFT, no \methodshort  (12B-pt)& .336 (5) & .192 (3) & \textbf{.129 (1)} & \textbf{.243 (1)} \\
\textit{Model Size ablations (no train)} & & & & \\
\quad 1B-pt (no SFT, no \methodshort) & .341 (6) & .219 (5) & .308 (4) & .429 (3) \\
\quad 4B-pt (no SFT, no \methodshort) & .351 (7) & .201 (4) & \underline{.174 (2)} & \underline{.314 (2)} \\
\quad 12B-pt (no SFT, no \methodshort) & .336 (5) & .192 (3) & \textbf{.129 (1)} & \textbf{.243 (1)} \\
\quad 27B-pt (no SFT, no \methodshort) & .312 (4) & \underline{.176 (2)} & \textbf{.120 (1)} & \textbf{.246 (1)} \\
\midrule
\rowcolor{gray!20} \textsc{Soft Task} & MP & CSC & Par & VEN \\
\rowcolor{gray!20} & (Manh. dist. $\downarrow$) & (Wass. dist. $\downarrow$) & (Wass. dist. $\downarrow$) & (Manh. dist. $\downarrow$) \\
\midrule
Ours (\methodshort + SFT + Demographics + ICL) & \textbf{.422 (1)} & \textbf{.746 (1)} & \textbf{1.20 (1)} & \textbf{.449 (1)} \\
\textit{Prompt ablations} & & & & \\
\quad no demographics & \underline{.430 (2)} & \textbf{.751 (1)} & \textbf{1.17 (1)} & \textbf{.458 (1)} \\
\quad no many-shot ICL (one-example)          & .448 (3) & \underline{.851 (2)} & 2.27 (3) & \textbf{.484 (1)} \\
\textit{Training Ablations} & & & & \\
\quad no SFT          & .486 (5) & .963 (3) & \textbf{1.15 (1)} & \textbf{.446 (1)} \\
\quad no \methodshort          & .450 (3) & \textbf{.749 (1)} & \textbf{1.21 (1)} & \textbf{.418 (1)} \\
\quad no SFT, no \methodshort
(12B-pt) & .507 (6) & .959 (3) & \textbf{1.21 (1)} & \textbf{.427 (1)} \\
\textit{Model Size ablations (no train)} & & & & \\
\quad 1B-pt (no SFT, no \methodshort) & .511 (7) & 1.13 (5) & 3.24 (4) & .703 (3) \\
\quad 4B-pt (no SFT, no \methodshort) & .526 (8) & 1.03 (4) & \underline{1.75 (2)} & \underline{.519 (2)} \\
\quad 12B-pt (no SFT, no \methodshort) & .507 (6) & .959 (3) & \textbf{1.21 (1)} & \textbf{.427 (1)} \\
\quad 27B-pt (no SFT, no \methodshort) & .462 (4) & \underline{.875 (2)} & \textbf{1.11 (1)} & \textbf{.413 (1)} \\
\end{tabular}
\caption{Ablation study results for a hypothetical competition between all entries shown, with the rank in parentheses. First place is bolded, second place is underlined.
Ties are determined sequentially by a two-sided Wilcoxon signed-rank test on item-level scores, failing to reject a difference with the rank leader above $\alpha=.05$ significance,
as in the actual competition (see Appendix \ref{app:ties} for details).
To see the results presented visually, also see Fig. \ref{fig:bothresults}.
}
\label{tab:ablation}
\end{table*}

\subsection{\method: Post-Training for In-Context Steerability}
Given this prompt structure, we post-train a language model on a large collection of $>40$ datasets involving human variation, epistemic uncertainty, or stochasticity, as described in \citet{sorensenmethod}. The post-training technique consists of unifying the datasets into the common description/input/output format,
removing any local dependencies by shuffling the in-context examples,
and fine-tuning with cross-entropy loss \textit{only} on the output/\texttt{<end\_of\_turn>} tokens (a.k.a., the highlighted tokens in the example Par prompt).
This post-training is meant to enhance the models' in-context learning abilities, teach the model to focus on predicting only the output tokens wrapped in the scaffolding, and improve calibration.
For additional details, please refer to \citet{sorensenmethod} and App. \ref{app:datasets}.

\subsection{Dataset-Specific Training}
Once we have the post-trained ICL model, we specialize the model to the particular dataset on which we plan to do inference. We do so by templatizing the entire train dataset in our prompt format, where all ratings in a given context are from the same rater, and performing additional supervised fine-tuning with cross-entropy loss on \textit{just} the output tokens (same format and loss as \methodshort, just with data only from target inference dataset). On MP/CSC, we include one training sequence per annotator, and on Par/VEN, which only have four annotators each, we batch into groups of 20 (Par)/30 (VEN) ratings per prompt and train on multiple sequences per annotator.

This could be seen in a way as meta-learning for the specific dataset \cite{vanschoren2018metalearningsurvey, min2022metaicllearninglearncontext}, with each rater being a different ``task" to which the model has to adapt in-context.

\subsection{In-Context Inference}
With the dataset-specific specialized model, we then do inference for each test instance / rater pair 1) by adding randomly-selected train examples into context until we hit a maximum token budget and 2) putting the target test instance at the end. We then directly calculate the model's probability of each label given the rater prompt, which is tractable due 
to there only being a small set of possible outputs.

Since MP and CSC's possible outputs all differ by only the initial token, only one forward pass per test rating was required. However, VEN and Par's outputs span multiple tokens, and thus required multiple forward passes in order to estimate the entire output probability distribution. Finally, we normalize the probability distribution to sum to one, removing probability mass on any token sequences that do not result in a valid label.

At the end, we have a probability estimate for all possible outputs for each test rater/instance combination.

\subsection{From Probabilities To Submission}
Up until this point we have taken a wholly perspectivist approach to predicting (a distribution over) how each rater will respond to each test instance. However, the perspectivist task requires a single answer candidate, and the soft task requires an distributional estimate of the entire population of raters will rate an instance.

For the perspectivist task, we submit the single response that minimizes the corresponding evaluation loss. For the two binary datasets, we submit the argmax response.\footnote{With the added constraint for VEN that each rater submits at least one positive annotation from entailment, neutral, and contradiction.}
For the two Likert datasets, we make the assumption that our label distribution estimate is well-calibrated, and submit the 50th percentile (median) Likert response of the distribution as this minimizes the expected absolute distance given draws from our distribution estimate.

For the soft task, the optimal distribution to submit under the evaluation criteria (Manhattan/Wasserstein) depends on
how well-calibrated our probability estimates are.
Here, rather than assuming a well-calibrated distribution, we experiment with a few approaches and submit the one that has the best dev set performance, which are as follows:
MP/Par: the averaged distributions for all test raters who annotated the instance; CSC/VEN: an equal average of 1) the averaged distributions and 2) the averaged perspectivist single-answer submissions.

\section{Results}
\label{sec:results}
We now outline how our system performed compared to others in the competition. Then, we ablate the components of our system to determine the effect of each on task performance.

\subsection{LeWiDi Competition Results}
\paragraph{Our system was the overall winner on both tasks.}
The final results can be seen in Table \ref{tab:results}.
For MP and CSC, our system had the lowest (best) scores for both the perspectivist and the soft tasks. We tied for second across the perspectivist evaluations for Par and VEN, tied for first on Par (soft), and got third on VEN (soft). Our average rank for the perspectivist and soft tasks was 1.5/1.5, which was the lowest overall rank across all teams, meaning our system was the overall winner for both the perspectivist and soft tasks.

\begin{figure*}[htbp]
\centering
\includegraphics[width=0.9\textwidth]{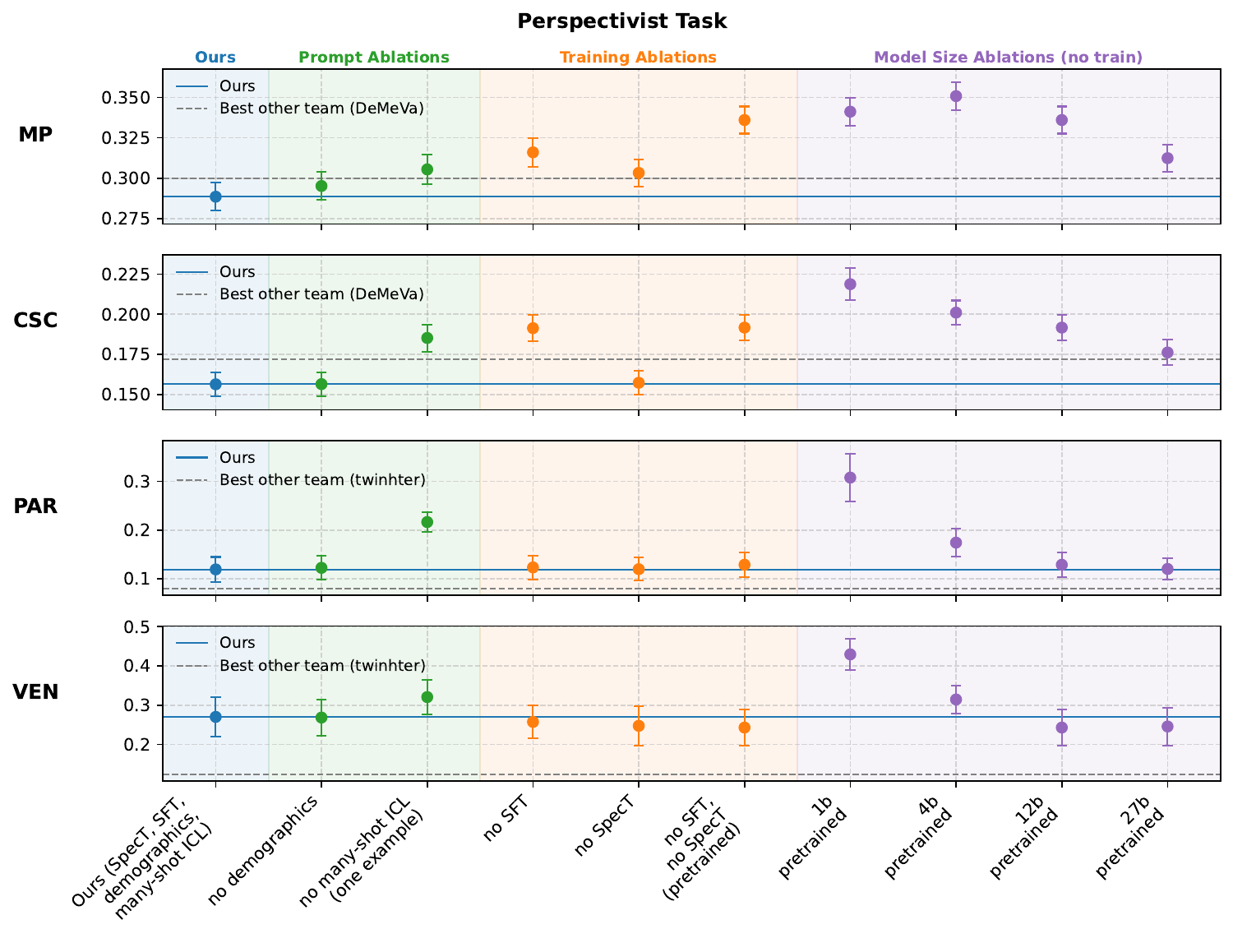}
\includegraphics[width=0.9\textwidth]{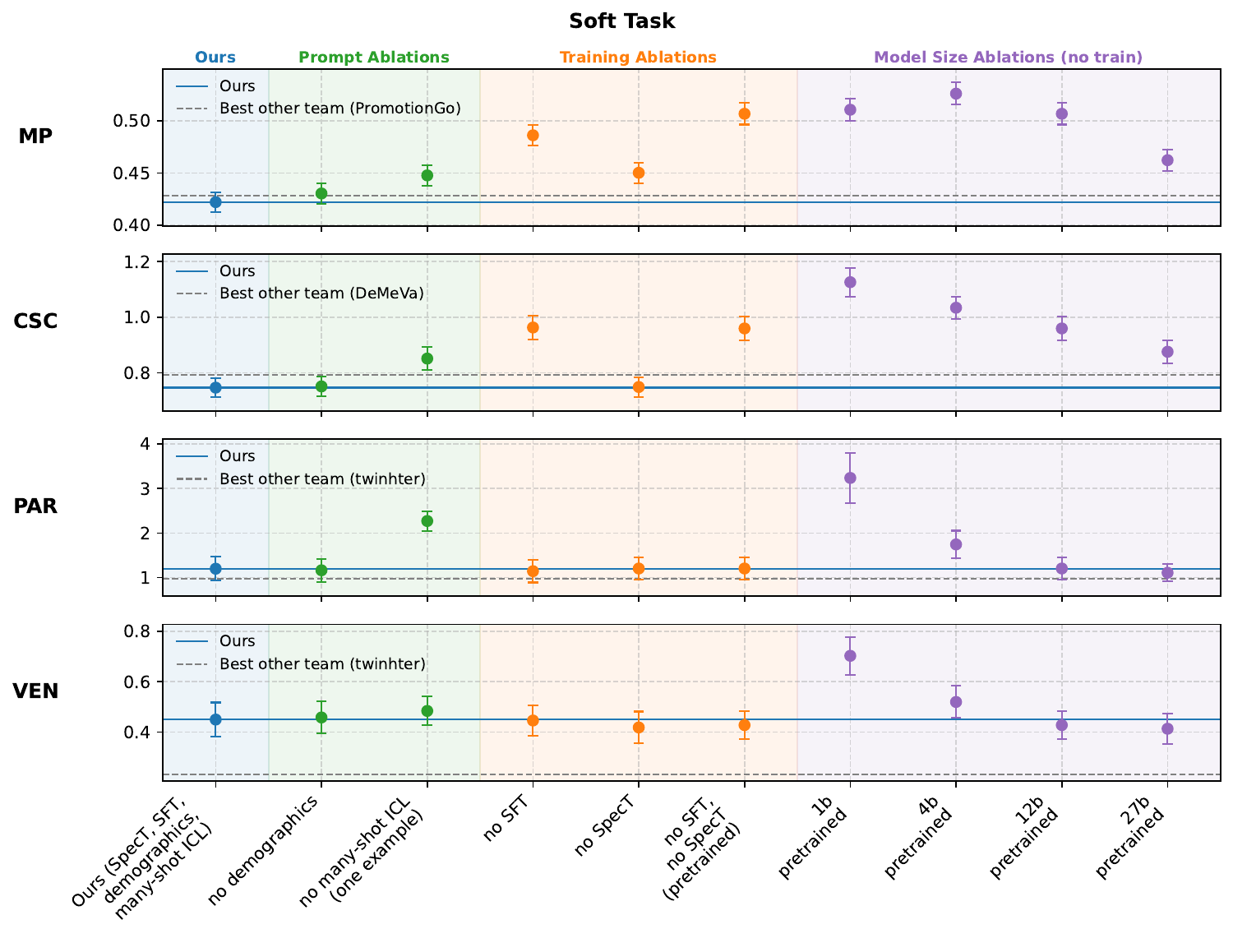}
\small{
\caption{
Ablation study results.
Perspectivist Task: For MP/VEN, error rate is reported, and for CSC/Par, absolute distance is reported (lower is better for both).
Soft Task: For MP/VEN, Manhattan distance is reported, and for CSC/Par, Wasserstein distance is reported (lower is better for both). 
Error bars indicate 95\% confidence intervals, computed as $\pm$ 1.96 times the standard error of the mean of instance-level scores.
Our system performance is shown as a solid line, and the best competing team performance is shown as a dashed line.
}
\label{fig:bothresults}
}
\end{figure*}

\subsection{System Ablations}

What was the effect of each component of our system? To answer this, we ablate 1) the continued model training via gradient descent, 2) the prompt components, and 3) the size of the underlying LLM.
We ablate the components and report the raw scores along with the rankings of a hypothetical competition between the ablated systems.
Results can be found in Table \ref{tab:ablation} and
Figures \ref{fig:bothresults}.

As a note, the MP and CSC datasets were much larger (3.8k/704 test instances) than the Par/VEN datasets (50/50 test instances). This allows us to make more confident comparisons for the MP/CSC results and affects the size of the available training data for model training.

\paragraph{In-context rater examples were crucial.}
In the inference prompts, we included many demonstration ratings per annotator (average: 16/29/35/29 across MP/CSC/Par/VEN, c.f. Table \ref{tab:prompt_stats}). To ablate the effect of the examples, we experimented with only including a single rater demonstration. Across all dataset/task combinations, we saw a substantial performance degradation when restricting to only one example (statistically significant across 7/8 comparisons).
This suggests that our system relies heavily upon the inclusion of these in-context demonstrations and the models' in-context learning ability.

Interestingly enough, this is true even for the Par dataset, where we include the annotator ID in the demographic description.\footnote{Due to an oversight that was not realized until after the conclusion of the competition, annotator ID was not included in the prompt for the other datasets.} Even though the model theoretically should be able to connect the annotator instances from its training data to that annotator through the annotator ID, performance substantially dropped when omitting the in-context examples (perspectivist: .119$\to$.216, soft: 1.20$\to$2.27). In other words, in our case, the model is much better able to leverage rater examples when provided concretely in-context at inference time, as opposed to relying on its ``soup" of model weights updated via gradient descent.

\paragraph{Demographics did not significantly help.}
Omitting the rater demographics, on the other hand, did not cause a significant drop in performance on CSC/Par/VEN, and caused only a slight drop in performance on MP. This suggests that the system was not able to significantly leverage sociodemographics in order to improve predictivity, in line with prior work \cite{orlikowski2025demographicsfinetuninglargelanguage, sorensen2025valueprofilesencodinghuman}.

\paragraph{Dataset-specific fine-tuning was important for the large datasets.}
For MP and CSC, omitting dataset-specific fine-tuning caused a significant drop in performance on both the perspectivist and soft tasks.
We hypothesize that this dataset-specific fine-tuning helped mainly due to 
1) (meta-)learning patterns of how to utilize in-context examples;
2) building better priors over how the average rater approaches the task;
and 3) specializing to the instance data distribution.

Dataset-specific fine-tuning did not, however, make a significant difference on Par/VEN. 
We hypothesize that the difference in result is largely due to dataset size, with only 400/388 annotations for Par/VEN in the training data. We also used the same hyperparameters for all datasets, and did not particularly adapt them to squeeze more out of the smaller dataset. Further optimization may be able to extract more signal, but machine learning systems generally struggle more in this low-data regime.

\paragraph{\method significantly helped on MP.} Applying \methodshort did significantly help on both MP tasks (perspectivist: .303$\to$.289, soft: .450$\to$.422), but did not significantly help or hurt on the other datasets.
We are not sure why it significantly helped in some cases and did not others, but it is not due to any additional irony detection training data, as that was not included in the \methodshort training mix (see Appendix \ref{app:datasets}).

\paragraph{Performance improves with model size, but size alone does not compensate for dataset-specific training.}
Due to computational constraints, we did not replicate our entire system (with \methodshort/SFT) across multiple model sizes. However, we did evaluate the pretrained models of the \texttt{gemma-3} model family (1B/4B/12B/27B) on which our 12B system was based in order to get a feel for the importance of model size. In general, we observe the expected trend that bigger is better. However, there does seem to be a particular jump in performance from 1B to 4B. Additionally, on the larger datasets where dataset-specific SFT helped (MP/CSC), our 12B system outperforms the 27B system without \methodshort/SFT.

\section{Discussion and Conclusion}
\label{sec:discussion}

In summary, our system was able to perform strongly across the board and was \textbf{the overall winner on both tasks}. However, it did perform particularly well (1st) on MP and CSC, which had many unique annotators and larger training datasets, and performed less well on Par/VEN (perspectivist: 2nd on Par, 2nd on VEN; soft: 1st on Par, 3rd on VEN), which had only four annotators each and much smaller training sets.

Our approach has many advantages, including: 1) a single model for each dataset, 2) potential adaptation at test time to new raters; 3) strong performance even in the limited data regime; 4) no dataset-specific assumptions; 5) same system for perspectivist and soft tasks.
However, some limitations include expensive inference
(see App \ref{app:inferencedetails}),\footnote{Although, this could be further optimized with techniques such as prompt caching \cite{gim2024promptcachemodularattention}.}
as prompt lengths are quite long in order to contain in-context rater examples 
and that the method is unable to effectively leverage additional rater demonstrations that do not fit in the context window.

In our ablation study, we found that in-context demonstrations are crucial for performance, dataset-specific tuning helps given enough data,
\method helped on MP,
and performance improves with model size (but scale alone does not make up for dataset-specific training).

Some interesting directions for future work include:
1) how performance scales with the number of in-context rater examples (including going beyond 3,000-token prompts),
2) whether selecting particular in-context examples at inference can outperform random selection,
3) the effect of including rater explanations on performance,
4) how well the approach generalizes to free-text / non-categorical tasks,
and 5) methods to better extract dataset-specific signal from smaller datasets (e.g., Par/VEN).

\section*{Acknowledgments}
This research was supported in part by DARPA under
the ITM program (FA8650-23-C-7316).

\newpage


\bibliography{custom}

\clearpage

\appendix

\section{Implementation details}
\label{app:implementation}
\subsection{General Details}
\begin{itemize}
    \item  All experiments were carried out using 1-4 80GB A100s.
    \item  For all experiments, since the \texttt{gemma-3-pt} models \cite{gemmateam2025gemma3technicalreport} do not have a trained embedding for \texttt{<start\_of\_turn>/<end\_of\_turn>}, we copy over the (un/)embedding weights for these tokens from the \texttt{gemma-3-it} models, as in \cite{sorensenmethod}.
    \item  Our \methodshort model is an early version of the model from \cite{sorensenmethod}. For more details, see App. \ref{app:datasets}.
\end{itemize}

\subsection{Dataset-specific SFT hyperparameters}
\begin{itemize}
    \item Training hardware: 4 80GB A100s
    \item \texttt{max\_length: 1024}
    \item \texttt{per\_device\_train\_batch\_size: 1}
    \item \texttt{gradient\_accumulation\_steps: 4}
    \item \texttt{learning\_rate: 1e-6}
\end{itemize}

\subsection{Inference Details}
\label{app:inferencedetails}
All inference was done on a single 80GB A100. MP needed a single forward pass per test rating:
($p(\{\texttt{"0","1"}\})$,
CSC also needed a single forward pass,
($p(\{\texttt{"1", "2", "3", "4", "5", "6"}\})$,
Par required three forward passes,
($p(\{\texttt{" "," -"}\})$,
$p(\{\texttt{"1", "2", "3", "4", "5"}\}|\texttt{" -"})$,
$p(\{\texttt{"0", "1", "2", "3", "4", "5"}\}|\texttt{" "})$),
and VEN required four forward passes
$p(\{\texttt{"entailment","contradiction", "neutral"}\})$,
($p(\{\texttt{"entailment","contradiction", "neutral"}\})$,
$p(\{\texttt{" neutral", " contradiction", "\}"}\}|\texttt{"entailment"})$,
$p(\{\texttt{" entailment", " contradiction", "\}"}\}|\texttt{"neutral"})$,
$p(\{\texttt{" entailment", " neutral", "\}"}\}|\texttt{"contradiction"})$).

The approximate run time for each inference pass on the entire test set was:
\begin{itemize}
    \item MP: 23 hours, 30 minutes;
    \item CSC: 4 hours;
    \item Par: 11 minutes;
    \item VEN: 11 minutes;
\end{itemize}
This was not well optimized however, and could potentially be sped up with methods such as prompt caching \cite{gim2024promptcachemodularattention} or vLLM \cite{kwon2023efficient}.

\section{Tie calculation}
\label{app:ties}
For calculating ties/significance, we used the competition organizer's code for the Wilcoxon signed-rank test to compare entries, as follows:
"For each of the four datasets and tasks, to determine ranking, we compared each team to the leading system within a cluster using the Wilcoxon signed-rank test on item-level results from the test sets. Teams were compared sequentially to the leader, and as long as no statistically significant difference was observed, they were assigned the same rank. This process continued until a team showed statistically distinct performance, at which point a new rank was introduced." (quoted from the competition organizers \cite{lewidi})

\clearpage
\section{Prompts}
\label{app:prompts}

Here, we include example prompts for the four datasets. \\
\textbf{\texttt{lewidi\_csc\_sarcasm\_detection\_individual}}
\begin{lstlisting}[style=plaincode]
Given a conversational context and response, rate how sarcastic the response is on a 1-6 scale.
Annotator demographics: Gender: Female; Age: 26
{"context": "Steve has been going out non-stop for the past two months because he needs a distraction from his recent breakup.  You are worried that he might be becoming a bit too destructive.  Steve says, \"ugh, worst hangover yet. I feel like crap.\"", "response": "maybe try some selfcare", "lang": "en"}
<start_of_turn>(*@\hlc[hlAmber]{3<end\_of\_turn>}@*)
{"context": "You and Steve have long been planning to go to a new bar in town. But, he has canceled on you three times without telling you why. And just now, he calls you and says, \"I'm so sorry, but I'm gonna have to bail again. Next time?\"", "response": "yeah let me know when you've made the plans", "lang": "en"}
<start_of_turn>(*@\hlc[hlGreen]{1<end\_of\_turn>}@*)
{"context": "Steve talks about the differences between two types of dinosaurs for an hour. You absolutely don't care about the topic.", "response": "anyways... next topic", "lang": "en"}
<start_of_turn>(*@\hlc[hlBlue]{1<end\_of\_turn>}@*)
{"context": "Steve borrowed your spare phone charger two months ago. Then he took your toaster a month ago. He did not return any of them.  And now, Steve says, \"can I borrow your suitcase? I need one for my trip next week.\"", "response": "not really, i think I'm going to need it on the weekend", "lang": "en"}
<start_of_turn>(*@\hlc[hlLavender]{3<end\_of\_turn>}@*)
{"context": "Steve bought a really expensive pair of shoes as a treat to himself for having finished a big project at work. The shoes go very well with his outfit today.", "response": "nice shoes!", "lang": "en"}
<start_of_turn>(*@\hlc[hlRed]{1<end\_of\_turn>}@*)
{"context": "Steve recently changed jobs. He is annoyed because he needs to deal with some bureaucracy regarding his health insurance. He says, \"I should have just stayed at my old job. If it hadn't been for this new job, I wouldn't have had to deal with so much crap.\"", "response": "maybe that's something you should've researched before but potentially ask for help or spend some time actually figuring this out.", "lang": "en"}
...
\end{lstlisting}

\textbf{\texttt{lewidi\_mp\_irony\_detection\_individual}}

\begin{lstlisting}[style=plaincode]
Given a post-reply pair from social media (Twitter/Reddit), determine whether the reply is ironic given the post. Context includes platform source, reply depth level, language variety, and language code. Binary irony detection task.
Annotator demographics:
{"post": "My company have basically said we can work from home if we feel safer doing so... but only with our direct manager's approval.\nBut no one has the stones to make the first move on my team. Plenty of other teams have people at home now. But my team get the vibe our manager would be a bit shit if we started.\nHonestly I would definitely feel safer. I can work 100% remote, and my office is  giant open plan with nearly 1000 people who are constantly travelling for work, so if this actually kicks off it'll be a fair nightmare for spreading.", "reply": "Just bite the bullet and ask better safe than sorry worst they can say is no.", "source": "reddit", "level": "1.0", "language_variety": "ie", "lang": "en"}
<start_of_turn>(*@\hlc[hlAmber]{0<end\_of\_turn>}@*)
{"post": "I\u2019ve heard it all now. Albanese has described himself as being \u201cEconomically Literate\u201d.", "reply": "@USER Of course he is. Don't forget he said he was an economic adviser to Bob Hawke. Trouble is Bob didn't know that and neither did anybody else.", "source": "twitter", "level": "1.0", "language_variety": "au", "lang": "en"}
<start_of_turn>(*@\hlc[hlGreen]{1<end\_of\_turn>}@*)
{"post": "Bit worried about it actually. Work in health care and I have asthma.   If I do get it. I am going to be as sick as anything.", "reply": "Fingers crossed you don't! I work in retail and surrounded by people who decide that shopping is the best idea when suffering with colds and sickness bugs. A bit like the health care sector cos I worked there too!", "source": "reddit", "level": "1.0", "language_variety": "gb", "lang": "en"}
<start_of_turn>(*@\hlc[hlBlue]{0<end\_of\_turn>}@*)
{"post": "Can't get it without being anti-national.", "reply": "Nah , everyone will get it", "source": "reddit", "level": "1.0", "language_variety": "in", "lang": "en"}
...
\end{lstlisting}

\textbf{\texttt{lewidi\_par\_paraphrase\_detection\_individual}}
\begin{lstlisting}[style=plaincode]
Given a pair of questions from Quora Question Pairs (QQP), assign a Likert scale score from -5 to 5 indicating how strongly the questions are paraphrases of one another, and provide a short explanation for your score.
Annotator demographics: annotator_id: Ann1; Gender: Male; Age: 26; Nationality: Chinese; Education: master student
{"question1": "What are some things new employees should know going into their first day at Exact Sciences?", "question2": "What are some things new employees should know going into their first day at Garmin?", "lang": "en"}
<start_of_turn>(*@\hlc[hlAmber]{\{"paraphrase\_rating": -1, "explanation": "The companies are different."\}<end\_of\_turn>}@*)
{"question1": "Who are the everyday heroes and heroines of life?", "question2": "What was everyday life like under Nazi rule?", "lang": "en"}
<start_of_turn>(*@\hlc[hlGreen]{\{"paraphrase\_rating": -5, "explanation": "Q1 asks about everyday heroes and heroines. Q2 is aobut ever
yday life under nazi rule"\}<end\_of\_turn>}@*)
{"question1": "What does 'sandiaga' mean?", "question2": "What does \u064a\u0639\u0646\u064a mean?", "lang": "en"}
<start_of_turn>(*@\hlc[hlBlue]{\{"paraphrase\_rating": -1, "explanation": "The words to be translated are different."\}<end\_of\_turn>}@*)
{"question1": "What is the best way to become a voracious reader?", "question2": "What is a voracious reader?", "lang": "en"}
<start_of_turn>(*@\hlc[hlLavender]{\{"paraphrase\_rating": -3, "explanation": "Q1 is about how to become a voracious reader. Q2 is about what is a voracious reader."\}<end\_of\_turn>}@*)
{"question1": "How do I believe in myself more?", "question2": "How can I believe in myself?", "lang": "en"}
<start_of_turn>(*@\hlc[hlRed]{\{"paraphrase\_rating": 3, "explanation": "The person in Q2 might not believe in themselves at all. The person in Q1 might have already believed in themselves."\}<end\_of\_turn>}@*)
{"question1": "What is does \"get swoll\" mean and where does it stem from?", "question2": "Where did leafy get his name from?", "lang": "en"}
...
\end{lstlisting}

\textbf{\texttt{lewidi\_varierrnli\_nli\_detection\_individual}}
\begin{lstlisting}[style=plaincode]
Given a premise and hypothesis from MNLI corpus, assign one or more labels from {Entailment, Neutral, Contradiction} indicating the logical relationship between them, and provide an explanation for your reasoning.
Annotator demographics: Gender: Female; Age: 22; Nationality: Chinese; Education: master student
{"context": "Even if auditors do not follow such other standards and methodologies, they may still serve as a useful source of guidance to auditors in planning their work under GAGAS.", "statement": "GAGAS requires strict compliance for auditors to follow.", "lang": "en"}
<start_of_turn>(*@\hlc[hlAmber]{\{"nli\_label": "entailment", "explanation": "The context suggests that auditors need to plan their work under GAGAS."\}<end\_of\_turn>}@*)
{"context": "In May 1967, Gallup found that the number of people who said they intensely disliked RFK--who was also probably more intensely liked than any other practicing politician--was twice as high as the number who intensely disliked Johnson, the architect of the increasingly unpopular war in Vietnam.", "statement": "Due to his attitudes on cheesecake, RFK was more disliked than Johnson.", "lang": "en"}
<start_of_turn>(*@\hlc[hlGreen]{\{"nli\_label": "neutral", "explanation": "The reason why RFK was more disliked than Johnson is not mentioned in the context."\}<end\_of\_turn>}@*)
{"context": "It was made up to look as much like an old-fashioned steam train as possible.", "statement": "It was built in the modern era to look like something built in the past.", "lang": "en"}
<start_of_turn>(*@\hlc[hlBlue]{\{"nli\_label": "entailment", "explanation": "The context mentions the building of an old-fashioned train, the word old-fashioned would only be used in the modern era. So the statement is true."\}<end\_of\_turn>}@*)
{"context": "Today it is possible to buy cheap papyrus printed with gaudy Egyptian scenes in almost every souvenir shop in the country, but some of the most authentic are sold at The Pharaonic Village in Cairo where the papyrus is grown, processed, and hand-painted on site.", "statement": "The Pharaonic Village in Cairo is the only place where one can buy authentic papyrus.", "lang": "en"}
...
\end{lstlisting}

\section{\methodshort Implementation}
\label{app:datasets}

The model used in our system was an early version of the model from \citet{sorensenmethod}.
The differences between our submission version and the final model are 1) a slightly modified prompt structure (see examples for details), 2) a slightly smaller dataset mix (see App. \ref{app:datasets}), and 3) an earlier hyperparameter set.

Hyperparameters:
\begin{itemize}
    \item Training hardware: 4 80GB A100s
    \item \texttt{max\_length: 1024}
    \item \texttt{per\_device\_train\_batch\_size: 1}
    \item \texttt{gradient\_accumulation\_steps: 512}
    \item \texttt{learning\_rate: 3e-6}
\end{itemize}

Here is the subset of datasets from \citet{sorensenmethod} that were used in training our system:
\begin{lstlisting}[style=plaincode]
ambient_ambiguity_detection
ambient_disambiguation
ambient_interpretation_labels
ambient_linguist_annotations
ambient_premise_hypothesis
babynames
bare_enron
bare_gsm8k
bare_hotpot
bare_lcb
binomial
cards
categorical
changemyview_categories
changemyview_posts
chatbotarena_assistant
chatbotarena_individual_prefs
chatbotarena_prompts
coinflip
dices
diffuse_distribution
flight
generativesocialchoice_freetext
generativesocialchoice_validation
geometric
geometric_beta
globaloqa
gsm8k_answer_from_question
gsm8k_question
gsm8k_question_answer
gsm8k_question_from_answer
habermas_categorical
habermas_individual
habermas_individual_categorical
habermas_opinions
habermas_question
haikus
hatespeech_comment
hatespeech_individual
helpsteer
hypergeometric
imdb
issuebench
jeopardy_answer_prediction
jeopardy_question_generation
multinomial
negative_binomial
netflix_individual_ratings
netflix_individual_views
newsgroups
normal
novacomet_hypothesis
novacomet_premise
numbergame_individual
numbergame_perc
opinionqa_individual
opinionqa_questions
polis_comment
polis_vote
poisson
popquorn_individual
popquorn_og_categorical
prism_prompts
prism_prompts_individual
pubmed
titanic_all_variables
titanic_survival_prediction
valueconsistency
valueprism_misc
valueprism_situation
valueprism_vrd
valueprism_vrds_noncontextual
wvs_individual
zipfian
\end{lstlisting}

\end{document}